\documentclass[letterpaper, 10 pt, conference]{llncs}  
\usepackage{amsmath}
\usepackage{array}
\usepackage{booktabs}
\usepackage{flushend}
\usepackage{graphicx}
\usepackage{multirow}
\usepackage{hyperref}
\usepackage{url}
\usepackage{wrapfig}
\usepackage{xcolor}
\usepackage{algorithm}
\usepackage{algpseudocode}
\usepackage{tabularx}
\newcolumntype{P}[1]{>{\centering\arraybackslash}p{#1}}

\title{\LARGE \bf
    Adapting Deep Visuomotor Representations with Weak Pairwise Constraints }

\author{Eric Tzeng\thanks{Authors contributed equally.}\inst{1}, Coline Devin$^{\star}$\inst{1}, Judy Hoffman\inst{1}, Chelsea Finn\inst{1},\\
  Pieter Abbeel\inst{1}, Sergey Levine\inst{1}, Kate Saenko\inst{2}, Trevor Darrell\inst{1}}
\institute{University of California, Berkeley \and Boston University}

\newcommand{\xInd}[1]{x^{(#1)}}

\newcommand{\poseGT}{\phi}

\newcommand{\loss}{\mathcal{L}}
\newcommand{\lPose}{\loss_\phi}

\newcommand{\lPairwise}{\loss_\text{pairwise}}
\newcommand{\lConf}{\loss_\text{conf}}

\newcommand{\weightConf}{\lambda}
\newcommand{\weightContrastive}{\nu}

\newcommand{\param}{\theta}
\newcommand{\paramD}{\param_D}
\newcommand{\paramRepr}{\param_\text{repr}}
\newcommand{\paramPose}{\param_\phi}
\newcommand{\paramCtrl}{\param_{\text{ctrl}}}

\begin{document}

\maketitle
\thispagestyle{empty}
\pagestyle{empty}

\begin{abstract}
Real-world robotics problems often occur in domains that differ significantly from the robot's
prior training environment. For many robotic control tasks, real world experience is expensive to 
obtain, but data is easy to collect in either an instrumented environment
or in simulation. We 
propose a novel domain adaptation approach for robot perception that adapts visual representations 
learned on a large easy-to-obtain source dataset (e.g. synthetic images) to a target real-world 
domain, without requiring expensive manual data annotation of real world data
before policy search. 
Supervised domain adaptation methods 
minimize cross-domain differences using pairs of aligned images that contain the same object or 
scene in both the source and target domains, thus learning a domain-invariant representation. 
However, they require manual alignment of such image pairs. Fully unsupervised adaptation methods 
rely on minimizing the discrepancy between the feature distributions across domains. We propose a 
novel, more powerful combination of both distribution and pairwise image alignment, and remove the 
requirement for expensive annotation by using weakly aligned pairs of images in the source and 
target domains. Focusing on adapting from simulation to real world data using a PR2 robot, we 
evaluate our approach on a manipulation task and show that by using weakly paired images, our 
method compensates for domain shift more effectively than previous techniques, enabling better 
robot performance in the real world.

\end{abstract}

\section{Introduction}
Transfer and domain shift are major challenges in learning-based robotic perception and control. Perception systems built using offline datasets often fail when deployed on a robot, robots trained to perceive and act in a laboratory setting might fail outside of the lab, and robots trained in simulation often fail in the real world. However,  accurate data annotations (such as the state of the world) are often only available in simulated or instrumented environments, which usually look too different from the real world to use directly. 
To enable adaptation of robotic perception between domains, we present a deep learning architecture that learns to map images from each domain into a common feature space.

  We propose a novel framework with losses for both pairwise alignment and distribution-level alignment. We also introduce a new algorithm for aligning source and target images without labels in the target domain.
This method is general and can be applied to many perception tasks, and we show that it increases performance on adapting pose estimation (predicting object keypoints) from synthetic images to real images.
 Furthermore, this technique can be used to pretrain visual features for visuomotor policy search. Recently proposed end-to-end visuomotor networks~\cite{levine15arxiv} can learn both image representations and the control policy for a particular task directly from visual data. In particular, the method in~\cite{levine15arxiv} first learns a convolutional network to predict keypoint locations from raw images, then fine-tunes the representation with guided policy search to map keypoints to actions. However, this previous method uses
 1000 pose annotated images to train the keypoint predictor.
We show that pretraining using our framework allows us to construct effective vision-based manipulation policies without any pose annotated real images.

Existing deep domain adaptation methods have focused on the category-level domain invariance task, and used optimization
 to generally reduce the discrepancy, or maximize confusion, between domains~\cite{tzeng15iccv,ganin15icml}; this is valuable, but misses a significant opportunity in the setting of synthetic to real image adaptation. 
It is often feasible to generate a large enough variety of synthetic images such that for each unlabeled real image, there exists a matching synthetic image.
This can provide instance level training constraints for a deep domain adaptation architecture that minimizes the distance between features of the instance pair.
Previous work has not tackled the problem of learning these pairs in settings where explicit annotations are unavailable.
Additionally, while such constraints have been explored in earlier adaptation schemes~\cite{saenko10eccv}, to our knowledge they have not been combined with contemporary deep discrepancy or deep confusion models.

\begin{figure}
\begin{center}
	\vspace{-0.5cm}
  \includegraphics[height=1.2in]{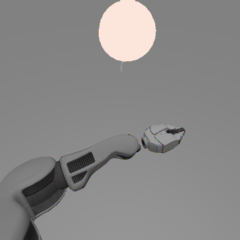}
  \quad
  \includegraphics[height=1.2in]{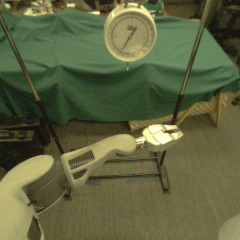}
\end{center}
\caption{
  A pair of corresponding synthetic (left) and real-world (right) images used for our pose estimation evaluation. Our method finds pairs without real-world supervision
}
\label{fig:arm_pair}
\end{figure}
\vspace{-0.5cm}
We report experiments with our framework on the pose pretraining stage of the visuomotor model of~\cite{levine15arxiv}, using a real and simulated PR2, as shown in Figure~\ref{fig:arm_pair}. We
also evaluate the learned representations by using them as input for training a visuomotor policy.
Our results confirm (1) there can be a significant domain shift in visuomotor task learning, (2) that domain adaptation methods specialized to the deep spatial feature point architecture introduced in~\cite{levine15arxiv} can learn to be relatively invariant to such shifts and improve performance, (3) that inclusion of pairwise constraints provides a performance boost relative to previous deep domain adaptation approaches based solely on  discrepancy minimization or domain confusion maximization, and (4) that, even in settings where pose annotations are unavailable for target domain imagery, annotations can be transferred from a source domain dataset (e.g. generated by a low-fidelity renderer).
We validate our method by training a visuomotor policy on the PR2 robot to perform a simple manipulation task.

\section{Related work}
In both vision and robotics, it has long been a desirable goal to use easily obtainable data (such as synthetic rendered images) to train models that are effective in real environments.
In robotics, past work has used domain adaptation and simulated data to reduce the need for labeled target domain examples. Lai and Fox used a variant of feature augmentation~\cite{daume07acl} to use human-made 3D models for  laser scan classification~\cite{lai2009RSS}.
Saxena et al. used rendered objects to learn to grasp from vision~\cite{saxena2008robotic}. 

Classically, in computer vision,
hand-engineered features were designed to be invariant to the domain shift between synthetic and real worlds, e.g., efforts dating from the earliest model alignment methods in computer vision using edge detection-based representations~\cite{brooks79ijcai}. One of the earliest visuomotor neural network learning methods, ALVINN~\cite{p-alvin-89}, exploited simulated training data of observed road shapes when training a multi-layer perceptron for an autonomous driving task. Many approaches to pose estimation in the recent decade were trained using rendered scenes from POSER and other human form rendering systems~\cite{shakhnarovich2003fast,urtasun08cvpr,taylor10nips}; reliance on fixed feature representations limited their performance, however, and state-of-the-art pose estimation methods generally train exclusively on real imagery~\cite{toshev13arxiv,tompson14nips}.

Traditional visual domain adaptation methods tackled the problem where a fixed representation extraction algorithm was used for both visual domains, and adaptation took the form of learning a transformation between the two spaces~\cite{saenko10eccv,gopalan11iccv,gong12cvpr} or regularizing the target domain model based on the source domain~\cite{yang07acmm,aytar11iccv}. Later
models improved upon this by proposing adaptation which both transformed the representation spaces and regularized the target model using the source data~\cite{duan12icml,hoffman13iclr}.
Since the resurgence in the popularity of convolutional networks for visual representation learning, adaptation approaches have been proposed to optimize the full target representation and model to better align with the source, for example by minimizing the maximum mean discrepancy~\cite{tzeng14arxiv,long15icml} or by minimizing the $a$-distance
(specific form of discrepancy distance~\cite{mansour14colt}) between the two distributions~\cite{ganin15icml,tzeng15iccv}.

Recently, a method has been proposed to use 3D object models to render synthetic training examples for training visual models with limited human annotations needed~\cite{sun14bmvc}. It was shown that there is a specific domain shift problem that arises when applying a synthetically trained visual model to the real world data. This paradigm of synthetic to real was further used to study deep representations and the types of invariances they learn by \cite{peng14iclr}.

While classic robotic perception already provides ample motivation for exploring scalable and effective domain adaptation methods, recent progress in deep reinforcement learning (RL) raises another intriguing possibility. Deep RL methods have shown remarkable performance and generality on tasks ranging from simulated locomotion to playing Atari games~\cite{mksga-padrl-13,slmja-trpo-15,lhphe-ccdrl-15}, but often at the cost of very high sample complexity. Other than the method in~\cite{levine15arxiv}, many of these methods are impractical to use directly on real physical systems due to the sample requirements, and a key question is whether policies learned with deep reinforcement learning in simulation could be extended for use in the real world. In this paper, we present an initial step in this direction by showing that vision systems trained on simulated data and adapted using our technique can be used to initialize deep visuomotor policies that achieve superior performance on real-world tasks, when compared to policies trained using small amounts of real-world data.

Previous attempts to learn transformations from source to target domains for visual domain adaptation such as \cite{kulis11cvpr} and \cite{saenko10eccv} have used a contrastive metric learning loss.
In these methods the learned adaptation was a kernelized transformation over a fixed representation. Earlier work introduced Siamese networks~\cite{bromley94nips,chopra05cvpr}, for which a shared representation is directly optimized using the contrastive loss for signature and face verification. These were later used for dimensionality reduction~\cite{hadsell06cvpr} and person hand and head pose alignment~\cite{taylor10nips}.
Taylor et al.~\cite{taylor10nips} further explored combining synthetic data along with real data to improve representation invariance and overall performance. However, this method used the synthetic data to regularize the learning of the real model and found that performance suffered once the amount of simulated data overwhelmed the amount of real world data. In contrast, our approach uses synthetic data to learn a complete model and uses a very limited number of real examples for refining and adapting that model. 

Recently, there has been considerable interest in learning visuomotor policies directly from visual imagery using deep networks \cite{rlv-arlrv-12,levine15arxiv,wsbr-etc-15,lhphe-ccdrl-15}. This tight coupling between perception and control simplifies both the vision and control aspects of the problem, but suffers from the major limitation that each new task requires collection, annotation, and training on real world visual data in order to successfully learn a policy. To overcome this issue, we explore how simulated imagery can be adapted for robotic tasks in the real world. Directly applying models learned in simulation to the real world typically does not succeed \cite{zhang15arxiv}, due to systematic discrepancies between real and simulated data. We demonstrate that our domain adaptation method can successfully perform pose estimation for a real robotic task using minimal real world data, suggesting that adaption from simulation to the real world can be effective for robotic learning.
In an earlier version of this paper \cite{tzeng15arxiv}, we demonstrated initial results using domain confusion constraints on PR2 visuomotor policies but without the pairwise constraint reported below. Contemporaneously to our work, \cite{Daftry2016ISER} also reported success with a domain confusion-style regularizer on a domain adaptive visual behavior task on autonomous MAV flight.

\section{Preliminaries}
We address the problem of adapting visual representations for robotic learning from a source domain where labeled data is easily accessible (such as simulation) to a target domain without labels. Domain adapation is often necessary because of \emph{domain shift}: a discrepancy in the data distributions between domains that prevents a model trained only on source data to perform well on target data. We define the problem as finding image features $f(x;\paramRepr)$ such that this representation allows learning visuomotor policies from a large dataset $x_S$ of labeled source images and a small dataset $x_T$ of unlabeled target images.

When training models for regression, we generally seek to take input images $x$ and directly output some label $\poseGT$.
This involves learning a representation $\paramRepr$ and a regressor $\paramPose$ that minimizes the following loss:
\begin{equation}
\label{eq:task}
  \lPose(x, \poseGT; \paramPose, \paramRepr) = \frac{1}{2K} \sum_{i=1}^K || \paramPose^Tf(\xInd{i}; \paramRepr) - \poseGT^{(i)} ||_2^2
\end{equation}
where $f(\xInd{i}; \paramRepr)$ denotes the feature vector corresponding to $\xInd{i}$ under the representation defined by $\paramRepr$.

However, collecting ground truth labels in the real world can be impractical, often requiring expensive instrumented setups.
As a result, it is difficult to gather enough training data to properly train models from scratch.
We instead rely on the existence of a simulator that can render synthetic versions of the task environment.
This enables us to quickly generate an unlimited amount of training data with full annotations by simply changing the environment configuration, recording the ground truth label, and rendering a view.

Ideally, we would be able to simply train on our rendered data and have the learned model transfer to the real world.
However, because they are acquired independently, our synthetic and real-world images differ significantly in appearance.
This discrepancy between the two domains is referred to as \emph{domain shift}, and generally results in reduced performance when attempting to directly transfer source models to the target domain.

To combat the negative effects of domain shift, we model this as a domain adaptation problem, with synthetic renders serving as our source domain, and  real-world images serving as our target domain.
We propose a model that augments the task loss with two additional adaptation loss functions designed to specifically align the two domains in feature space.
This ensures we learn a model that successfully performs the task and transfers robustly between domains.

\section{Domain alignment with weakly supervised pairwise constraints}\label{sec:gda}
\begin{figure*}
  \centering
  \includegraphics[width=.70\linewidth]{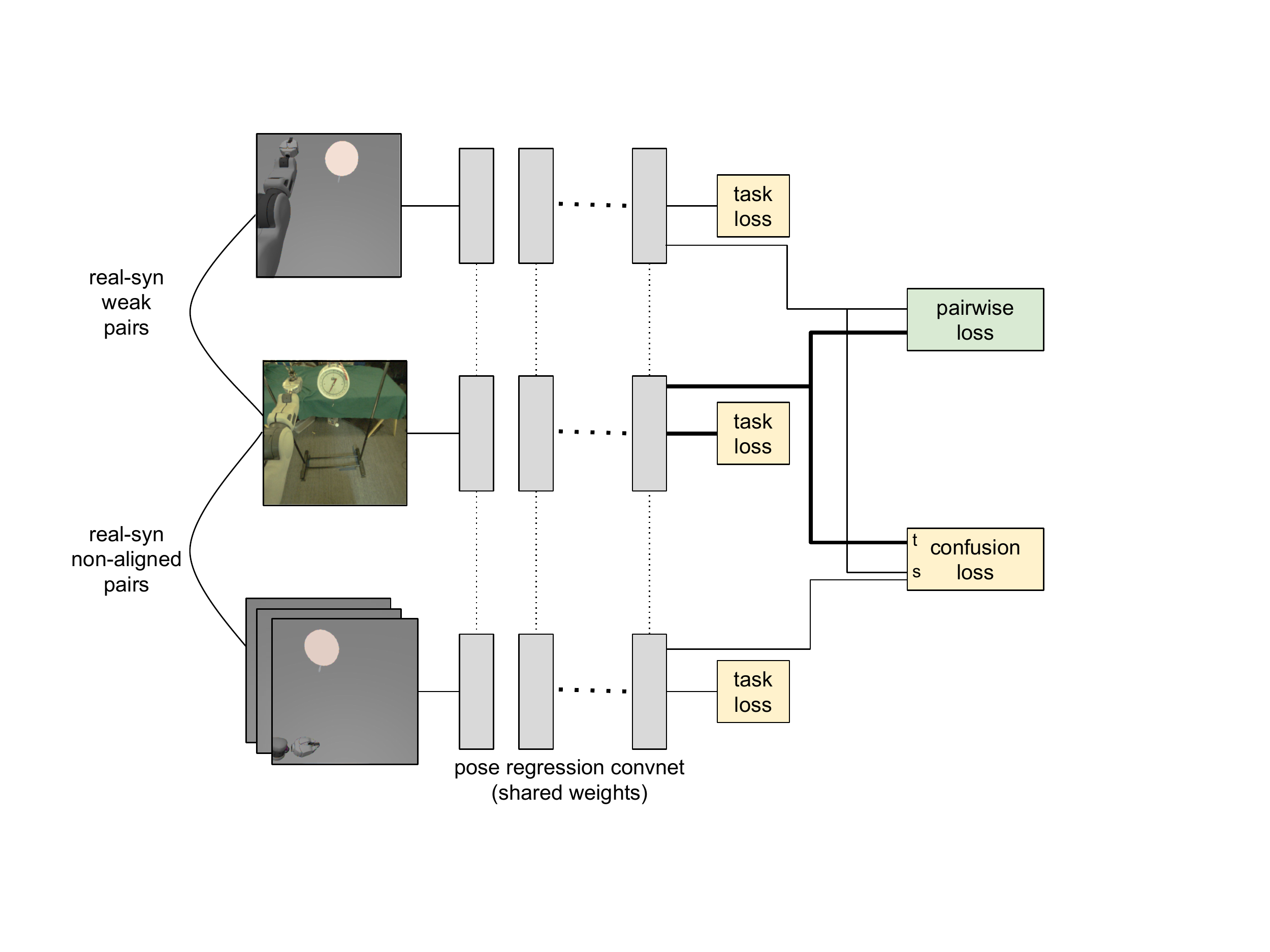}
  \caption{
    After determining a weak pairing between source and target images, optimization proceeds via backpropagation on our model architecture.
    Our model combines a task loss, a domain confusion loss for aligning domains at the distribution level, and a pairwise loss for aligning specific pairs of source and target images.
    Together, these three losses ensure that our model learns to accurately perform the task while remaining robust to domain shift.}
  \label{fig:arch}
\end{figure*}
Our method attempts to solve the domain shift problem via two approaches.
The first is a distribution-based approach, which seeks to align the source domain with the target domain in feature space.
By ensuring that all images lie in the same general neighborhood in representation space, we better facilitate the transfer of task-relevant features from source to target.
The second approach incorporates weakly supervised pairwise constraints, and seeks to ensure that images with identical labels are treated identically by the network, regardless of their originating domain.
This encourages the network to disregard domain-specific features in favor of features that are relevant to the perception task.
Together with the task loss, these approaches ensure that we learn a representation that is meaningful to the chosen visual task while remaining robust to the source-target domain shift.

\textbf{Domain confusion loss.}
To align the source and target domains at the overall distribution level, we adopt the domain confusion loss introduced by \cite{tzeng15iccv,ganin15icml}.
The model trains a domain classifier $\paramD$ that attempts to correctly classify each image into the domain it originates from.
In parallel, the loss $\lConf$ tries to learn a representation $\paramRepr$ such that the domain classifier cannot distinguish the two domains in feature space.
This loss is the negative cross entropy loss between the predicted domain label of each image $x$ and a uniform distribution over the $D$ domains, which is minimized the domain classifier is maximally confused:
\begin{equation}
  \lConf(x_S, x_T, \paramD; \paramRepr) = 
  - \sum_{x \in (x_S \cup x_T)} \sum_d \frac{1}{D} \log q_d(x, \paramD, \paramRepr).
\end{equation}
Here, $q$ corresponds to the domain classifier activations:
\begin{equation}
  q(x, \paramD; \paramRepr) = \text{softmax}(\paramD^T f(x; \paramRepr))
\end{equation}

\textbf{Pairwise loss.}
While the confusion loss ensures that the source and target domains as a whole are treated similarly by the model, it does not make use of the task labels.
Thus, we include an additional term that seeks to find specific pairs of source and target images with similar labels and align them in representation space.
By explicitly aligning images with similar labels, we can optimize the representation to focus only on task-relevant features.
However, we assume that task labels are unavailable in the target domain.
Thus, we need to determine a pairing $P$ of the target images $x_T$ with the target images $x_S$
so that we can ensure that their distances in the feature space defined by $\paramRepr$ lie close together.
We write this objective as the loss function
\begin{equation}
  \lPairwise(x_S, x_T; P, \paramRepr) = \sum_{(i, j) \in P} \left[ \frac{1}{2} \rho\left(\xInd{i}_S, \xInd{j}_T; \paramRepr\right)^2 \right], \\
\end{equation}
where we define our distance function $\rho$ as the Euclidean distance in the feature space corresponding to $\paramRepr$:
\begin{equation}
  \rho\left(\xInd{i}_S, x^{(j)}_T; \paramRepr\right) = \left\lVert f(\xInd{i}_S; \paramRepr) - f(\xInd{j}_T; \paramRepr) \right\rVert_2.
\end{equation}
Intuitively, this objective encourages a pairing $P$ that correctly matches target and source images, as well as a representation $\paramRepr$ that is task-sensitive while disregarding domain-specific features.
However, because the source-target pairing $P$ and the feature representation $\paramRepr$ depend on each other, it is not immediately clear how to directly optimize for both simultaneously.
Thus, we propose an iterative approach.

First, we minimize $\lPairwise$ with respect to the source-target pairing $P$.
We begin by finding an initial representation $\paramRepr$ that minimizes the task loss $\lPose$ and optionally $\lConf$ on only the source imagery.
Once this source-only model has been trained, we extract a feature representation for every image in our dataset, both source and target.
These representations are used to find a source image nearest-neighbor for each target image, thereby determining a weak pairing $P$.
Finding such a pairing additionally enables us to transfer task labels between each pair of images, thus annotating the target images using the labels from their corresponding source images.
These transferred weak labels can then be used to minimize the task loss $\lPose$ over the target images as well.

Once $P$ has been determined, we keep it fixed and minimize $\lPairwise$ with respect to the representation $\paramRepr$ to ensure that pairs lie close in feature space.
We note that when used to optimize $\paramRepr$, this loss function is similar to the contrastive loss function introduced by \cite{hadsell06cvpr}. 
As typically formulated, the contrastive loss function seeks to draw paired images closer together in feature space while pushing unpaired images apart.
However, our source dataset has many examples similar to any particular target image, which means there are often many other valid source-target pairs in the dataset that are not explicitly identified.
The dissimilarity term in the contrastive loss function would force these unlabeled similar pairs apart, making the optimization poorly conditioned, so our pairwise loss omits this dissimilarity term.

\textbf{Complete objective.}
Our full model thus minimizes the joint loss function
\begin{equation}
\label{eq:completeloss}
\begin{split}
  \loss(x_S, &\poseGT_S, x_T, \poseGT_T, P, \paramD; \paramPose, \paramRepr) = \\
  &\lPose(x_S, \poseGT_S; \paramPose, \paramRepr) + \lPose(x_T, \poseGT_T; \paramPose, \paramRepr) \\
  & + \weightConf\lConf(x_S, x_T, \paramD; \paramRepr) \\
  & + \weightContrastive\lPairwise(x_S, x_T; P, \paramRepr)
\end{split}
\end{equation}
where the hyperparameters $\weightConf$ and $\weightContrastive$ trade off how strongly we enforce domain confusion and weakly supervised pairwise constraints.

The feature used to form $P$ is a low-level convolutional feature of a network trained to perform the visual task on the source data.
In this feature space, we match each target image with its nearest neighbor in the source domain.
Because the feature used to determine $P$ is from a network trained to perform the perception task, it focuses primarily on task-relevant features of the image.
After the pairing $P$ has been determined, we can then minimize the complete loss function outlined in Equation~\ref{eq:completeloss} via backpropagation.
This procedure of determining a weak alignment $P$ and using it to learn a domain-invariant representation is summarized in Algorithm~\ref{algorithm}.

We depict the architecture setup for a given sampled target image in Figure~\ref{fig:arch}.
The task loss is applied to all images the network sees, regardless of whether they came from the source or target environment.
Because the target examples do not have labels, we use the labels transferred from the source using the pairing $P$.
Each pair is input to the pairwise loss which pushes the feature representations of the explicitly paired images closer together.
Finally, all images are additionally optimized by the confusion loss, which seeks to make the representation agnostic to the overall differences between the two domains.

The combination of losses presented here is architecture-agnostic, thereby making our method applicable to many different visual tasks.
We implement our networks using the Caffe framework~\cite{caffe}, and plan to release the code and datasets from our experiments upon acceptance of this paper.
\begin{algorithm}
\caption{Learning domain-invariant image features}\label{algorithm}
\begin{algorithmic}[1]
\State Collect $x_S$ source domain images with labeled object pose
\State Collect $x_T$ target domain images
\State Minimize $\lPose(x_S, \poseGT_S; \paramPose, \paramRepr) + \weightConf\lConf(x_S, x_T, \paramD; \paramRepr)$ with respect to $\paramPose, \paramRepr$
\For{$x_T^{(j)}$ in $x_T$}
\State $i^* = \text{arg}\min_i ||f_\text{conv1}(x_S^{(i)};\paramRepr) - f_\text{conv1}(x_T^{(j)};\paramRepr)||_2$
\State Add $(i^*, j)$ to $P$
\EndFor 
\State Minimize $\loss(x_S, \poseGT_S, x_T, \poseGT_T, P, \paramD; \paramPose, \paramRepr)$ with respect to $\paramPose, \paramRepr$
\end{algorithmic}
\end{algorithm}

\section{Adapting visuomotor control policies }

As mentioned above, our domain adaptation approach is general and can be applied to many visual tasks.
Here we  use it to directly adapt deep visual representations for pose estimation and visual policy learning. We build upon the end-to-end architecture presented by \cite{levine15arxiv}
for training deep visuomotor policies that can learn to accomplish tasks such as screwing a cap onto a bottle or
placing a coat hanger on a rack. The method first pretrains a convolutional
neural network on a pose estimation task, then finetunes this network with guided
policy search to map from input image to action.  Guided policy search is initialized
with trajectories from a fully observed state (where the locations of both the
manipulated and target object are known), but once learned, the policy only requires visual
input at test time.

Once we have learned a visual representation that is robust to the synthetic-real domain shift and can effectively locate salient objects in a scene, we use guided policy search (GPS)
with these features to train a parametrized controller $\paramCtrl$.
GPS turns reinforcement learning into a supervised learning problem by using time-varying linear
controllers to collect (observation, control) data that is used to train a neural network policy. During training, the position of the target object is known, but the neural network policy is trained to act based on the visual feature points; at test time, this policy can succeed solely from vision without being provided the location of the target.

Like in~\cite{ftddl-lvfsr-15}, we fit time-varying linear models to the robot joint angles and velocities and use these to collect a dataset of feature points, feature point velocities, joint angles, and joint efforts. We use this dataset to train a neural network policy $\paramCtrl$. The feature points are generated by the $\paramRepr$ trained with our method, and we do not backpropagate gradients from $\paramCtrl$ through $\paramRepr$ during policy learning.
As in~\cite{levine15arxiv},
we used BADMM to jointly optimize the controllers and neural network with a penalty
on the KL divergence between them. $\paramCtrl$ is 2 layer network with 40 hidden units per layer
that takes the
learned feature points and joint state as input and outputs joint efforts. 
Unlike in~\cite{ftddl-lvfsr-15}, we do not apply any filtering or smoothing to the feature points. We refer the reader to~\cite{levine15arxiv} for a more in depth explanation of the BADMM GPS algorithm.
The final result is a visuomotor control policy from images features pretrained solely on unannotated real imagery and low-fidelity synthetic renderings, while the policy itself is trained in the real world.

We empirically evaluate our method in a variety of experimental settings.
We begin with an evaluation on a simple pose estimation task in Section~\ref{sec:robot_pose}.
Next, we investigate the quality of synthetic-real pairings produced by our unsupervised alignment method.
Finally, we use the learned pairings to train a representation via our method, then use this representation to train a full visuomotor control policy on a ``hook loop'' manipulation task in Section~\ref{sec:hook_task}.
These experiments demonstrate the effectiveness of incorporating synthetic imagery into the pretraining of visuomotor policies.

\subsection{Supervised robotic pose estimation evaluation}
\label{sec:robot_pose}
As a self-contained evaluation of our visual adaptation method, we first evaluate our method in a supervised setting, using a pose estimation task that is representative of the visual estimation required for robotic visuomotor control. This is intended as a toy task to evaluate the use of known pairs for simulation to real world adaptation. By using the gripper, we are able to generate images that are exactly paired between the domains. 

We first obtain real world images with gripper pose annotations using the PR2's forward kinematics.
We also collected pose labeled images from the Gazebo simulator, where we know the exact location of all objects, and we can specifically obtain paired images by replaying the joint angles used in the real world data collection. With this data, we train a model to regress to the 3D gripper pose from an image.
We adopt the deep spatial feature point architecture introduced by Levine et al.~\cite{levine15arxiv}.
Both the domain confusion loss ($\weightConf=0.1$) and pairwise loss ($\weightContrastive=0.01$) are applied at the third convolutional layer, after the ReLU nonlinearity.
As before, when both losses are employed simultaneously, we further halve each of their weights.
Results from this experimental setting are presented in Table~\ref{table:pr2_pose}.

\begin{table}
\centering
\caption{
  Using pairwise constraints improves pose estimation.
  We report supervised evaluation results averaged over 3 trials on PR2 gripper pose estimation using 5 labeled and paired real examples.
  Each real example is paired with a corresponding synthetic image.
  Minibatches are sampled such that an equal number of real and synthetic images are present.
  We report the average error of the prediction in centimeters.
  We find that, through combining both a domain confusion loss and a pair alignment loss, we are able to improve performance by 20\% (relative).
}
\begin{tabular}{lccc}\toprule
\textbf{Method}                                    &
\textbf{\#Sim}&
\textbf{\#Real}&
\textbf{Error (cm)} \\ \midrule
Synthetic only &1005 &0                              & $25.37 \pm 1.18$                       \\
Real only      &0 &5                             &  $4.43 \pm 0.23$                        \\
Synthetic and real   & 1005&5                        &  $7.74 \pm 3.90$                        \\ \midrule
Domain confusion~\cite{tzeng15iccv}  & 1005&0      &  $6.68 \pm 0.01$                        \\
Pairwise    loss      &1005&5                      &  $5.21 \pm 2.48$                        \\
Domain alignment with strong pairwise constraints &1005 &5 &  $3.98 \pm 0.02$                \\ \midrule
Oracle             &0&1000                     &  $0.90 \pm 0.13$                         \\ \bottomrule
\end{tabular}
\label{table:pr2_pose}
\end{table}

The results indicate that adaptation with paired examples yields improved performance.
We find that incorporating synthetic imagery during training is nontrivial, confirming our hypothesis that simulation to real world has a significant domain shift.
Simply combining synthetic and real imagery into one large training set negatively impacts performance, due to slight variations in appearance and viewpoint.
We see that domain confusion alone does not help either, since domain confusion does not offer a way to learn the specific viewpoint variations between the real and synthetic domains.
Nonetheless, by exploiting the presence of pairs, our method is able to account for these differences, performing better than all other baselines.
Comparing against the ``Oracle" setting, in which we train on 1000 labeled real examples, we see that our method is able to remove most of the negative effects of domain shift despite training on relatively few real examples.
(For additional results on vision-only adaptation from CAD models to real PASCAL images, we refer the reader to our earier report  \cite{tzeng15arxiv}.)

\subsection{Unsupervised synthetic-real alignment evaluation}
\label{sec:alignment_eval}
\begin{wrapfigure}[18]{R}{4.4cm}\begin{center}
  \includegraphics[width=\linewidth]{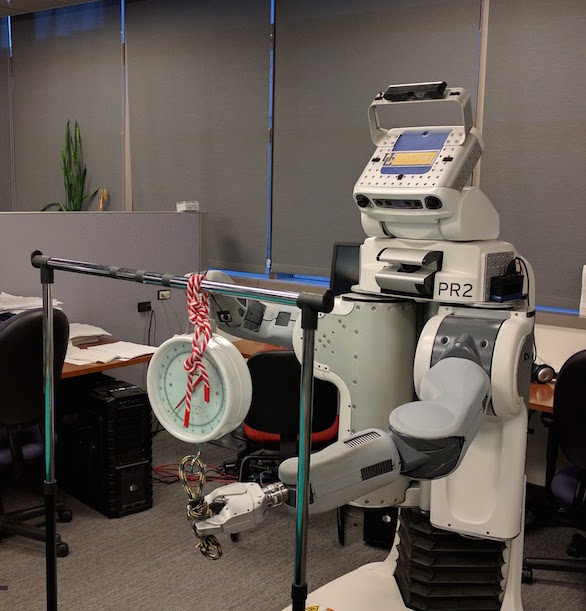}
\end{center}\vspace{-0.2cm}
\caption{
  In the ``hook loop'' task, the PR2 must position a loop of rope over the hook of a supermarket scale.
}
\label{fig:hook_task}
\end{wrapfigure}

To evaluate the effectiveness of our alignment method, we transfer pose annotations from paired synthetic images to their corresponding real images, then compute the error relative to the real-world ground truth pose annotations.
In order to test on a real control setting, we perform this experiment on the ``hook loop'' task introduced in~\cite{ftddl-lvfsr-15}, where the robot is expected to place a loop of rope on a hook, as depicted in Figure~\ref{fig:hook_task}.
We generate low-fidelity renderings of the PR2 and a hook in 4000 different configurations and attempt to align these with 100 real-world images of the task without hook pose annotations. As the goal is to learn a policy that can place the loop on an arbitrarily located hook, the policy must locate the hook from visual input.

\begin{table}
\centering
\caption{
  Comparing the pairing error for different strategies of learning $f_\text{conv1}$ for weak alignment.
  We compare using only the task loss during pretraining against combining both the task loss and the domain confusion images and report the average error between the object positions within each pair.
  We see that both the task loss as well as the task loss with confusion do significantly better than random, and in simpler settings their performance approaches that of the optimal alignment (reported as Oracle) if the real labels were known.
}
\begin{tabularx}{0.66\textwidth}{lcc}
\multicolumn{3}{c}{\textbf{Error of hook pose in weak pairings (cm)}}\\\toprule
\textbf{Method}  & \textbf{Static camera} &  \textbf{Head motion} \\ \midrule
Random pairs            & $22.7 \pm 0.4$  & $23.9 \pm 0.6$ \\
Task loss               & $5.9 \pm 0.2$   & $10.9 \pm 2.0$ \\
Task loss + confusion   & $6.1 \pm 0.4$   & $10.6 \pm 2.0$ \\
\midrule
Oracle (known real labels)                  & $4.1$           & $4.9$ \\ \bottomrule
\end{tabularx}
\label{table:alignment_error}
\end{table}

To learn the representation used for producing the alignment in this setting, we attempt to estimate the 3D pose of the target hook.
We evaluate both the alignment produced using the simple synthetic-only model, as well as a model trained with an additional domain confusion loss.
Table~\ref{table:alignment_error} shows the resuls of this experiment on two experimental settings: one with a fixed camera, and one in which the head of the robot (and the camera as well) moves around slightly.
The relatively low error in the results indicates that the alignments are generally of high quality.
\begin{figure*}[t]
  \centering
  \includegraphics[width=0.8\linewidth]{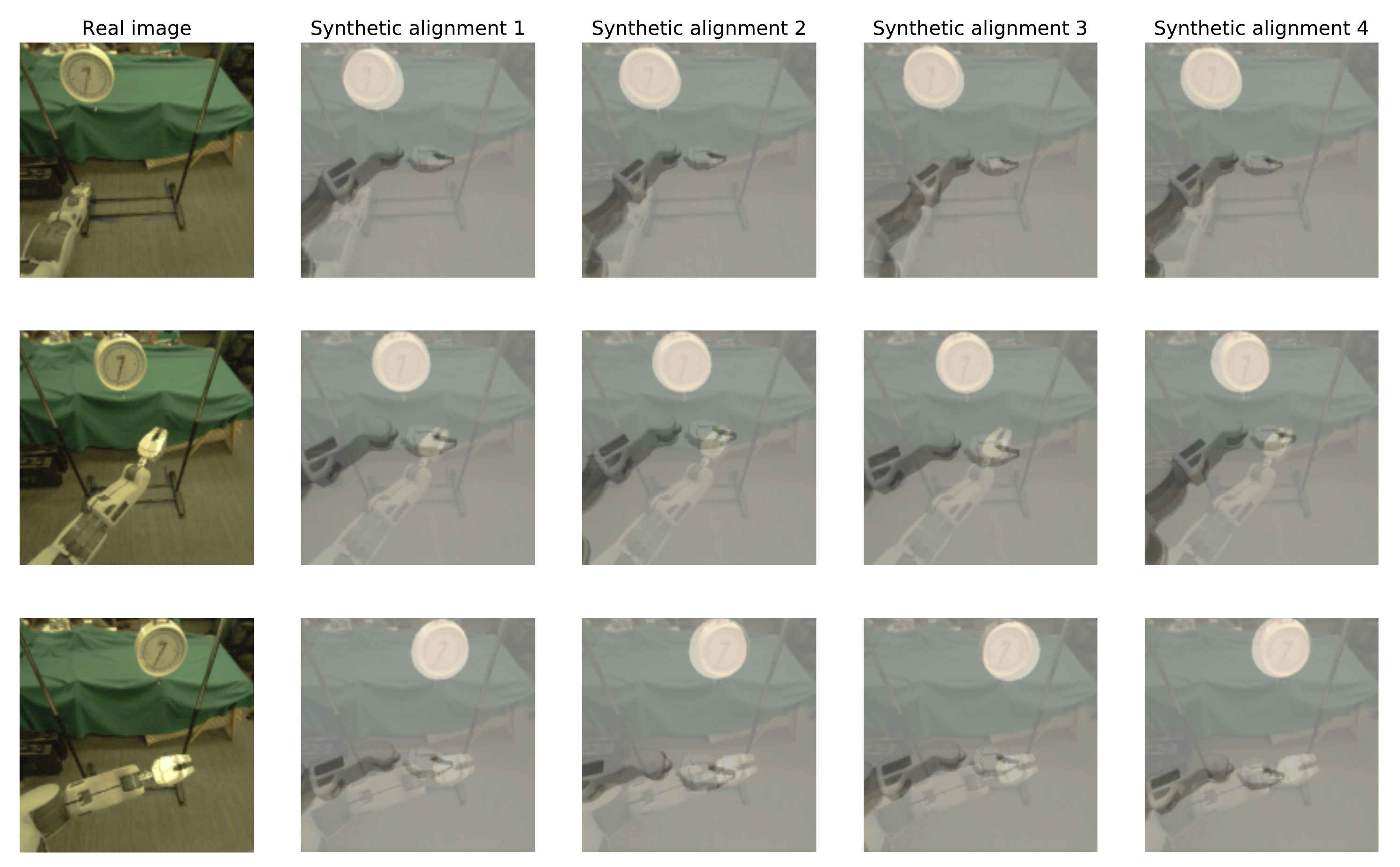}
  \caption{
    Example alignments generated by our unsupervised synthetic-real alignment method in the static camera setting.
    The first column shows an example real image, and the next four columns show the top four corresponding images from our rendered dataset. \textbf{The goal is to match the hook position, with the arm position being irrelevant}, because the policy needs to be conditioned on the hook position.
    We overlay a translucent version the real image on the synthetic images to better show the quality of our alignment.
          }
  \label{fig:alignment}
\end{figure*}

Visual inspection of the results also indicates that our method produces high-quality pairings.
Figure~\ref{fig:alignment} shows example results of our unsupervised alignment method in the static camera setting using the representation trained only on the synthetic data.
The hooks in the synthetic renderings match quite closely with the hooks in the corresponding real images.
As expected, the position of the arm is largely ignored as desired, and the alignment focuses primarily on the portion of the image that is relevant for the pose estimation task.

\subsection{Visuomotor policies for manipulation tasks}
\label{sec:hook_task}
After determining the synthetic-real pairings using our method, we retrain the pose predictor on the
combined data to learn the final feature points $\paramRepr$.
To evaluate these feature points, we
set up the ``hook loop'' task from~\cite{ftddl-lvfsr-15}. This task requires a PR2 to
bring a loop of rope to the hook of a supermarket scale, as depicted in Figure~\ref{fig:hook_task}. As the location of the scale
is not instrumented, the robot must adjust its actions by visually perceiving the
location of the hook/scale. 

We used four target hook positions along a bar to learn the linear dynamics and
generate trajectories. GPS
was run for 13 iterations, where each iteration obtained 5 sample trajectories for 4 training hook position. The linear-quadratic controller was given
only the arm joint state, while the neural network policy was given the arm joint state
as well as the learned feature point $(x,y)$ positions and velocities.

The performance of the final policy $\paramCtrl$ was measured by testing it 14 times:
twice at each of 7 positions (including the 4 training positions). Success was defined
as the loop being on the hook. As shown in Table~\ref{table:hook_task},
the features learned with our method allowed GPS to learn a much more accurate policy than the other methods not using labeled real images.

We also compared against the deep spatial autoencoder
from \cite{ftddl-lvfsr-15}. Trained on either 100 or 500 images, this method did not perform well, as the feature points tended to model the robot arm's position rather than the hook. Without the simulated hook pose supervision that our method has, the network has no incentive to model the hook over the much more varied positions of the arm and gripper.
We also trained an ``Oracle'' controller. The feature points used were from a pose
estimation model trained directly on 500 real images with ground truth data. This controller performed equally well to the one trained with adapted features on only 100 unlabeled real images.

\begin{table}
\centering
\caption{
  Performance of visuomotor tasks trained using domain alignment with weakly supervised pairwise constraints.
  We report the percentage of successful attempts at placing a loop of rope on a hook after training with 12 iterations of GPS. Each experiment was repeated 3 times.
}
\begin{tabularx}{\textwidth}{Xccc}\toprule
\textbf{Method} & \textbf{\# Sim}\hspace{1em} & \textbf{\# Real (unlabeled)}&\textbf{Success rate} \\ \midrule
Synthetic only &4000& 0& 38.1\% $\pm$ 8\% \\
Autoencoder (100) &0 & 100& 28.6\% $\pm$ 25\%\\
Autoencoder (500) &0& 500& 33.2\% $\pm$15\%\\
Domain alignment with randomly assigned pairs
&4000& 100 &33.3\% $\pm$16\%\ \\
Domain alignment with weakly\newline supervised pairwise constraints & 4000& 100&\textbf{76.2\% $\pm$ 16\%} \\ \midrule
Oracle & 0& 500 (labeled)&71.4\% $\pm$ 14\% \\
\bottomrule
\end{tabularx}
\label{table:hook_task}
\end{table}

Because of the optimization that happens during guided policy search, the performance
of the final controller is dependent on the quality of the feature points that are
passed in: if the feature points give $\paramCtrl$ enough information about the
position of the hook, then the controller will learn to use it. However, if the
feature points are not consistent enough in where they activate (such as in
many of our baselines), the controller cannot learn a policy that takes the hook location
into account. For example, when the controller failed a trial
it put the loop at a possible hook position, but not at the current hook position.
These results show that we can successfully learn visual features that are sufficient for control from synthetic data
and a small number of unlabeled real images.

In contrast to our prior work, which required either ground truth pose labels for the real-world images~\cite{levine15arxiv} or fifty 100-frame videos for a total of 5000 images for unsupervised learning~\cite{ftddl-lvfsr-15}, our method only uses 100 unlabeled real-world images.
Being able to use unlabeled images is important for for practical real-world robotic applications, where determining the ground truth pose of movable objects in the world with a high degree of precision typically requires specialized equipment such as motion capture.

\section{Conclusion}

In this paper, we present a novel model for domain adaptation that is able to exploit the presence of weakly paired source-target examples.
Our model extends existing adaptation architectures by combining pairwise and distribution alignment loss functions, and optimizaing over weak label assignments. 
Because of its generality, our method is applicable to a wide variety of deep adaptation architectures and tasks. 
Through a pose estimation task, we experimentally validate the importance of using image pairs and show that they are integral to achieving strong adaptation performance.
We  demonstrate the ability to adapt  in settings where pose annotations on real-world data is unavailable.

We address domain adaptation for visual inputs in the context of robotic state estimation. The tasks used in our robotic evaluation involve estimating information that is highly relevant for robotic control \cite{Pastor_ICRA_2013}, as well as for pretraining visuomotor control policies \cite{levine15arxiv}. While we show successful transfer of simulated data for learning real-world visual tasks, training full control policies entirely in simulation will also require tackling the question of physical adaptation, to account for the mismatch between simulated and real-world physics. Addressing this question in future work would pave the way for large-scale training of robotic control policies in simulation.

\bibliography{references}

\begin{thebibliography}{10}
\providecommand{\url}[1]{#1}
\csname url@rmstyle\endcsname
\providecommand{\newblock}{\relax}
\providecommand{\bibinfo}[2]{#2}
\providecommand\BIBentrySTDinterwordspacing{\spaceskip=0pt\relax}
\providecommand\BIBentryALTinterwordstretchfactor{4}
\providecommand\BIBentryALTinterwordspacing{\spaceskip=\fontdimen2\font plus
\BIBentryALTinterwordstretchfactor\fontdimen3\font minus
  \fontdimen4\font\relax}
\providecommand\BIBforeignlanguage[2]{{%
\expandafter\ifx\csname l@#1\endcsname\relax
\typeout{** WARNING: IEEEtran.bst: No hyphenation pattern has been}%
\typeout{** loaded for the language `#1'. Using the pattern for}%
\typeout{** the default language instead.}%
\else
\language=\csname l@#1\endcsname
\fi
#2}}

\bibitem{levine15arxiv}
S.~Levine, C.~Finn, T.~Darrell, and P.~Abbeel, ``End-to-end training of deep
  visuomotor policies,'' \emph{Journal of Machine Learning Research}, vol.~17,
  2016.

\bibitem{tzeng15iccv}
E.~Tzeng, J.~Hoffman, T.~Darrell, and K.~Saenko, ``Simultaneous deep transfer
  across domains and tasks,'' in \emph{International Conference in Computer
  Vision (ICCV)}, 2015.

\bibitem{ganin15icml}
Y.~Ganin and V.~Lempitsky, ``Unsupervised domain adaptation by
  backpropagation,'' in \emph{International Conference in Machine Learning
  (ICML)}, 2015.

\bibitem{saenko10eccv}
K.~Saenko, B.~Kulis, M.~Fritz, and T.~Darrell, ``Adapting visual category
  models to new domains,'' in \emph{Proc. ECCV}, 2010.

\bibitem{daume07acl}
H.~D. III, ``Frustratingly easy domain adaptation,'' \emph{ACL}, vol.~45, pp.
  256–--263, 2007.

\bibitem{lai2009RSS}
K.~Lai and D.~Fox, ``3d laser scan classification using web data and domain
  adaptation.'' in \emph{Robotics: Science and Systems, 2009}, 2009.

\bibitem{saxena2008robotic}
A.~Saxena, J.~Driemeyer, and A.~Y. Ng, ``Robotic grasping of novel objects
  using vision,'' \emph{The International Journal of Robotics Research},
  vol.~27, no.~2, pp. 157--173, 2008.

\bibitem{brooks79ijcai}
R.~Brooks, R.~Greiner, and T.~Binford, ``The acronym model-based vision
  system,'' in \emph{International Joint Conference on Artificial Intelligence
  6}, 1979, pp. 105--113.

\bibitem{p-alvin-89}
D.~Pomerleau, ``{ALVINN:} an autonomous land vehicle in a neural network,'' in
  \emph{Advances in Neural Information Processing Systems (NIPS)}, 1989.

\bibitem{shakhnarovich2003fast}
G.~Shakhnarovich, P.~Viola, and T.~Darrell, ``Fast pose estimation with
  parameter-sensitive hashing,'' in \emph{Computer Vision, 2003. Proceedings.
  Ninth IEEE International Conference on}.\hskip 1em plus 0.5em minus
  0.4em\relax IEEE, 2003, pp. 750--757.

\bibitem{urtasun08cvpr}
R.~Urtasun and T.~Darrell, ``Sparse probabilistic regression for
  activity-independent human pose inference,'' in \emph{Computer Vision and
  Pattern Recognition, 2008. CVPR 2008. IEEE Conference on}, June 2008, pp.
  1--8.

\bibitem{taylor10nips}
G.~W. Taylor, R.~Fergus, G.~Williams, I.~Spiro, and C.~Bregler,
  ``Pose-sensitive embedding by nonlinear nca regression,'' in \emph{Advances
  in Neural Information Processing Systems 23}, J.~Lafferty, C.~Williams,
  J.~Shawe-Taylor, R.~Zemel, and A.~Culotta, Eds.\hskip 1em plus 0.5em minus
  0.4em\relax Curran Associates, Inc., 2010, pp. 2280--2288.

\bibitem{toshev13arxiv}
A.~Toshev and C.~Szegedy, ``Deeppose: Human pose estimation via deep neural
  networks,'' \emph{CoRR}, vol. abs/1312.4659, 2013.

\bibitem{tompson14nips}
J.~J. Tompson, A.~Jain, Y.~Lecun, and C.~Bregler, ``Joint training of a
  convolutional network and a graphical model for human pose estimation,'' in
  \emph{Advances in Neural Information Processing Systems 27}, Z.~Ghahramani,
  M.~Welling, C.~Cortes, N.~Lawrence, and K.~Weinberger, Eds.\hskip 1em plus
  0.5em minus 0.4em\relax Curran Associates, Inc., 2014, pp. 1799--1807.

\bibitem{gopalan11iccv}
R.~Gopalan, R.~Li, and R.~Chellappa, ``Domain adaptation for object
  recognition: An unsupervised approach,'' in \emph{Proc. ICCV}, 2011.

\bibitem{gong12cvpr}
B.~Gong, Y.~Shi, F.~Sha, and K.~Grauman, ``Geodesic flow kernel for
  unsupervised domain adaptation,'' in \emph{Proc. CVPR}, 2012.

\bibitem{yang07acmm}
J.~Yang, R.~Yan, and A.~G. Hauptmann, ``Cross-domain video concept detection
  using adaptive svms,'' \emph{ACM Multimedia}, 2007.

\bibitem{aytar11iccv}
Y.~Aytar and A.~Zisserman, ``Tabula rasa: Model transfer for object category
  detection,'' in \emph{IEEE International Conference on Computer Vision},
  2011.

\bibitem{duan12icml}
L.~Duan, D.~Xu, and I.~W. Tsang, ``Learning with augmented features for
  heterogeneous domain adaptation,'' in \emph{Proc. ICML}, 2012.

\bibitem{hoffman13iclr}
J.~Hoffman, E.~Rodner, J.~Donahue, K.~Saenko, and T.~Darrell, ``Efficient
  learning of domain-invariant image representations,'' in \emph{International
  Conference on Learning Representations}, 2013.

\bibitem{tzeng14arxiv}
E.~Tzeng, J.~Hoffman, N.~Zhang, K.~Saenko, and T.~Darrell, ``Deep domain
  confusion: Maximizing for domain invariance,'' \emph{CoRR}, vol.
  abs/1412.3474, 2014.

\bibitem{long15icml}
M.~Long, Y.~Cao, J.~Wang, and M.~I. Jordan, ``Learning transferable features
  with deep adaptation networks,'' in \emph{International Conference in Machine
  Learning (ICML)}, 2015.

\bibitem{mansour14colt}
Y.~Mansour, M.~Mohri, and A.~Rostamizadeh, ``Domain adaptation: Learning bounds
  and algorithms,'' in \emph{COLT}, 2009.

\bibitem{sun14bmvc}
B.~Sun and K.~Saenko, ``From virtual to reality: Fast adaptation of virtual
  object detectors to real domains,'' in \emph{British Machine Vision
  Conference (BMVC)}, 2014.

\bibitem{peng14iclr}
\BIBentryALTinterwordspacing
X.~Peng, B.~Sun, K.~Ali, and K.~Saenko, ``Exploring invariances in deep
  convolutional neural networks using synthetic images,'' \emph{CoRR}, vol.
  abs/1412.7122, 2014. [Online]. Available:
  \url{http://arxiv.org/abs/1412.7122}
\BIBentrySTDinterwordspacing

\bibitem{mksga-padrl-13}
V.~Mnih, K.~Kavukcuoglu, D.~Silver, A.~Graves, I.~Antonoglou, D.~Wierstra, and
  M.~Riedmiller, ``Playing {Atari} with deep reinforcement learning,''
  \emph{NIPS '13 Workshop on Deep Learning}, 2013.

\bibitem{slmja-trpo-15}
J.~Schulman, S.~Levine, P.~Moritz, M.~Jordan, and P.~Abbeel, ``Trust region
  policy optimization,'' in \emph{International Conference on Machine Learning
  (ICML)}, 2015.

\bibitem{lhphe-ccdrl-15}
T.~P. Lillicrap, J.~J. Hunt, A.~Pritzel, N.~Heess, T.~Erez, Y.~Tassa,
  D.~Silver, and D.~Wierstra, ``Continuous control with deep reinforcement
  learning,'' \emph{arXiv preprint arXiv:1509.02971}, 2015.

\bibitem{kulis11cvpr}
B.~Kulis, K.~Saenko, and T.~Darrell, ``What you saw is not what you get: Domain
  adaptation using asymmetric kernel transforms,'' in \emph{Proc. CVPR}, 2011.

\bibitem{bromley94nips}
J.~Bromley, I.~Guyon, Y.~LeCun, E.~S\"{a}ckinger, and R.~Shah, ``Signature
  verification using a ``siamese" time delay neural network,'' in
  \emph{Advances in Neural Information Processing Systems 6}, J.~Cowan,
  G.~Tesauro, and J.~Alspector, Eds.\hskip 1em plus 0.5em minus 0.4em\relax
  Morgan-Kaufmann, 1994, pp. 737--744.

\bibitem{chopra05cvpr}
S.~Chopra, R.~Hadsell, and Y.~LeCun, ``Learning a similarity metric
  discriminatively, with application to face verification,'' in \emph{Computer
  Vision and Pattern Recognition, 2005. CVPR 2005. IEEE Computer Society
  Conference on}, vol.~1.\hskip 1em plus 0.5em minus 0.4em\relax IEEE, 2005,
  pp. 539--546.

\bibitem{hadsell06cvpr}
R.~Hadsell, S.~Chopra, and Y.~LeCun, ``Dimensionality reduction by learning an
  invariant mapping,'' in \emph{Proc. Computer Vision and Pattern Recognition
  Conference (CVPR'06)}.\hskip 1em plus 0.5em minus 0.4em\relax IEEE Press,
  2006.

\bibitem{rlv-arlrv-12}
M.~Riedmiller, S.~Lange, and A.~Voigtlaender, ``Autonomous reinforcement
  learning on raw visual input data in a real world application,'' in
  \emph{International Joint Conference on Neural Networks}, 2012.

\bibitem{wsbr-etc-15}
M.~Watter, J.~Springenberg, J.~Boedecker, and M.~Riedmiller, ``Embed to
  control: a locally linear latent dynamics model for control from raw
  images,'' in \emph{Advances in Neural Information Processing Systems (NIPS)},
  2015.

\bibitem{zhang15arxiv}
F.~{Zhang}, J.~{Leitner}, M.~{Milford}, B.~{Upcroft}, and P.~{Corke},
  ``{Towards Vision-Based Deep Reinforcement Learning for Robotic Motion
  Control},'' \emph{ArXiv e-prints}, Nov. 2015.

\bibitem{tzeng15arxiv}
\BIBentryALTinterwordspacing
E.~Tzeng, C.~Devin, J.~Hoffman, C.~Finn, X.~Peng, S.~Levine, K.~Saenko, and
  T.~Darrell, ``Towards adapting deep visuomotor representations from simulated
  to real environments,'' \emph{CoRR}, vol. abs/1511.07111, 2015. [Online].
  Available: \url{http://arxiv.org/abs/1511.07111}
\BIBentrySTDinterwordspacing

\bibitem{Daftry2016ISER}
S.~Daftry, J.~A. Bagnell, and M.~Hebert, ``Learning transferable policies for
  monocular reactive mav control,'' in \emph{International Symposium on
  Experimental Robotics}, 2016.

\bibitem{caffe}
Y.~Jia, E.~Shelhamer, J.~Donahue, S.~Karayev, J.~Long, R.~Girshick,
  S.~Guadarrama, and T.~Darrell, ``Caffe: Convolutional architecture for fast
  feature embedding,'' \emph{arXiv preprint arXiv:1408.5093}, 2014.

\bibitem{ftddl-lvfsr-15}
C.~Finn, X.~Tan, Y.~Duan, T.~Darrell, S.~Levine, and P.~Abbeel, ``Deep spatial
  autoencoders for visuomotor learning,'' in \emph{International Conference on
  Robotics and Automation (ICRA)}, 2016.

\bibitem{Pastor_ICRA_2013}
P.~Pastor, M.~Kalakrishnan, J.~Binney, J.~Kelly, L.~Righetti, G.~Sukhatme, and
  S.~Schaal, ``Learning task error models for manipulation,'' in \emph{IEEE
  International Conference on Robotics and Automation}, 2013.

\end{thebibliography}
\bibliographystyle{IEEEtran}

\end{document}